\documentclass[11pt]{article}

\usepackage[final]{acl}

\usepackage{times}
\usepackage{latexsym}
\usepackage{enumitem}
\usepackage{amsfonts}
\usepackage{amsmath}
\usepackage{bm}
\usepackage{soul}
\usepackage{multirow}
\usepackage[T1]{fontenc}
\usepackage[normalem]{ulem}
\usepackage[utf8]{inputenc}
\usepackage{svg}

\usepackage{inconsolata}
\usepackage{enumitem}
\setlist{topsep=0pt, partopsep=0pt, itemsep=0pt, parsep=0pt}
\usepackage{graphicx}
\makeatletter
\g@addto@macro\normalsize{%
  \setlength\abovedisplayskip{6pt plus 2pt minus 2pt}%
  \setlength\belowdisplayskip{6pt plus 2pt minus 2pt}%
  \setlength\abovedisplayshortskip{4pt plus 1pt minus 1pt}%
  \setlength\belowdisplayshortskip{4pt plus 1pt minus 1pt}%
}
\makeatother
\title{RePrompT: Recurrent Prompt Tuning for Integrating Structured EHR Encoders with Large Language Models}

\author{ Arya Hadizadeh Moghaddam, Drew Ross \\
\textbf{ Mohsen Nayebi Kerdabadi, Dongjie Wang, Zijun Yao}\thanks{Corresponding author.} \\
University of Kansas, USA \\
\texttt{\{a.hadizadehm, drewross, mohsen.nayebi, wangdongjie, zyao\}@ku.edu}
}

\begin{document}
\maketitle
\begin{abstract}
Large Language Models (LLMs) have shown strong promise for mining Electronic Health Records (EHRs) by reasoning over longitudinal clinical information to capture context-rich patient trajectories. However, leveraging LLMs for structured EHRs (e.g., standardized diagnosis and medication codes) presents two key challenges.
First, translating time-stamped EHR sequences into plain text can obscure both temporal structure and code identities, weakening the ability to capture code co-occurrence and longitudinal regularities.
Second, unlike cohort-trained predictive models that learn a shared, task-aligned representation space across patients, LLMs are often applied in a case-isolated inference setting where each patient is processed independently without leveraging population-level patterns.
To address these challenges, we introduce \textbf{RePrompT}, a time-aware LLM framework that integrates structured EHR encoders through prompt tuning, without modifying underlying architectures.
Specifically, RePrompT recurrently incorporates latent states from prior visits to preserve longitudinal information, and injects population-level information through trainable prompt tokens derived from a cohort-trained, task-aligned EHR encoder.
Experiments on MIMIC-III and MIMIC-IV demonstrate that RePrompT consistently outperforms both EHR-based and LLM-based baselines across multiple clinical prediction tasks.

\end{abstract}

\section{Introduction}
Electronic Health Records (EHRs) capture comprehensive information on patients’ diagnoses, procedures, and treatments across longitudinal clinical visits, and provide context-rich trajectories that enable data-driven clinical decision support systems \cite{gct,jiang2023graphcare}. While Large Language Models (LLMs) \cite{normalllm2} have shown promising results in EHR mining tasks, such as mortality and readmission prediction \cite{healtllm1, healthllm2}, two significant challenges remain in effectively applying LLMs to structured EHR signals.

The first challenge arises from the difficulty LLMs face in capturing the temporal structure of EHRs when longitudinal data are linearized into plain text for input.
A patient’s history typically consists of multiple visits over time, and the evolving trajectory across these visits plays a critical role in determining downstream outcomes \cite{safedrug}. For example, a history of chronic kidney disease (CKD) increases the likelihood of subsequent comorbidities such as cardiovascular disease (CVD) appearing in later visits \cite{bozkurt2016contributory}.
However, converting sequential EHR into textual descriptions can obscure both temporal dependencies and discrete identities of clinical codes.
Although inserting explicit separators in prompts to verbally denote visit boundaries can weakly encode temporality \cite{tan2024language, liu2019hierarchical}, the model's ability to process structured EHR \cite{healthllm7} still remains insufficient.
A promising direction is to incorporate temporal awareness into prompt tuning by enabling the model to explicitly access latent states from prior visits. This design strengthens visit differentiation and supports modeling of longitudinal progression, while avoiding substantial modifications to the existing LLM architecture.

The second challenge \cite{healthllm3, llmhealth4, llmhealth5} lies in the limited ability of LLMs to explicitly leverage population-level and task-specific representations for prediction. In traditional cohort-trained approaches \cite{GRAM, hap, retain}, models are optimized on a population of patients for a predefined clinical outcome. 
As a result, a shared, task-aligned representation space across patients enables the discovery of meaningful patterns, such as disease co-occurrence, longitudinal progression, and ontology relationships that recur across peers for prediction support.
In contrast, LLMs typically encode EHR information in a general-purpose manner and perform inference for each patient in a case-isolated setting. 
Without a shared, task-aligned patient representation space, LLMs lack an explicit mechanism to aggregate information from other patients to support prediction for a given individual. 

A naive solution is to directly include similar patient profiles in the prompt, for example, via few-shot learning. However, the large scale and high dimensionality of modern EHR datasets make this approach impractical.
A more promising direction is to integrate the complementary strengths of cohort-trained EHR models into LLMs. However, simple post-hoc fusion of embeddings from independently trained models is often suboptimal, as the representations are not jointly aligned. Recent advances in prompt tuning \cite{lester2021power, wu2023infoprompt} provide an effective alternative. By introducing trainable prompt tokens grounded in representations learned from cohort-trained EHR encoders \cite{vu2021spot}, LLMs can be adapted to incorporate patient-shared embeddings alongside context-rich clinical reasoning, which enables more principled structured EHR modeling.

To this end, as illustrated in Figure \ref{llmVSLLMClin}, we propose \textbf{RePrompT}, an adaptable LLM-based predictor that integrates structured EHR encoders through \textbf{Re}current \textbf{Promp}t \textbf{T}uning. Our proposed approach makes the following contributions:

\begin{figure}[t]
    \centering
	\includegraphics[width= 8.0cm]{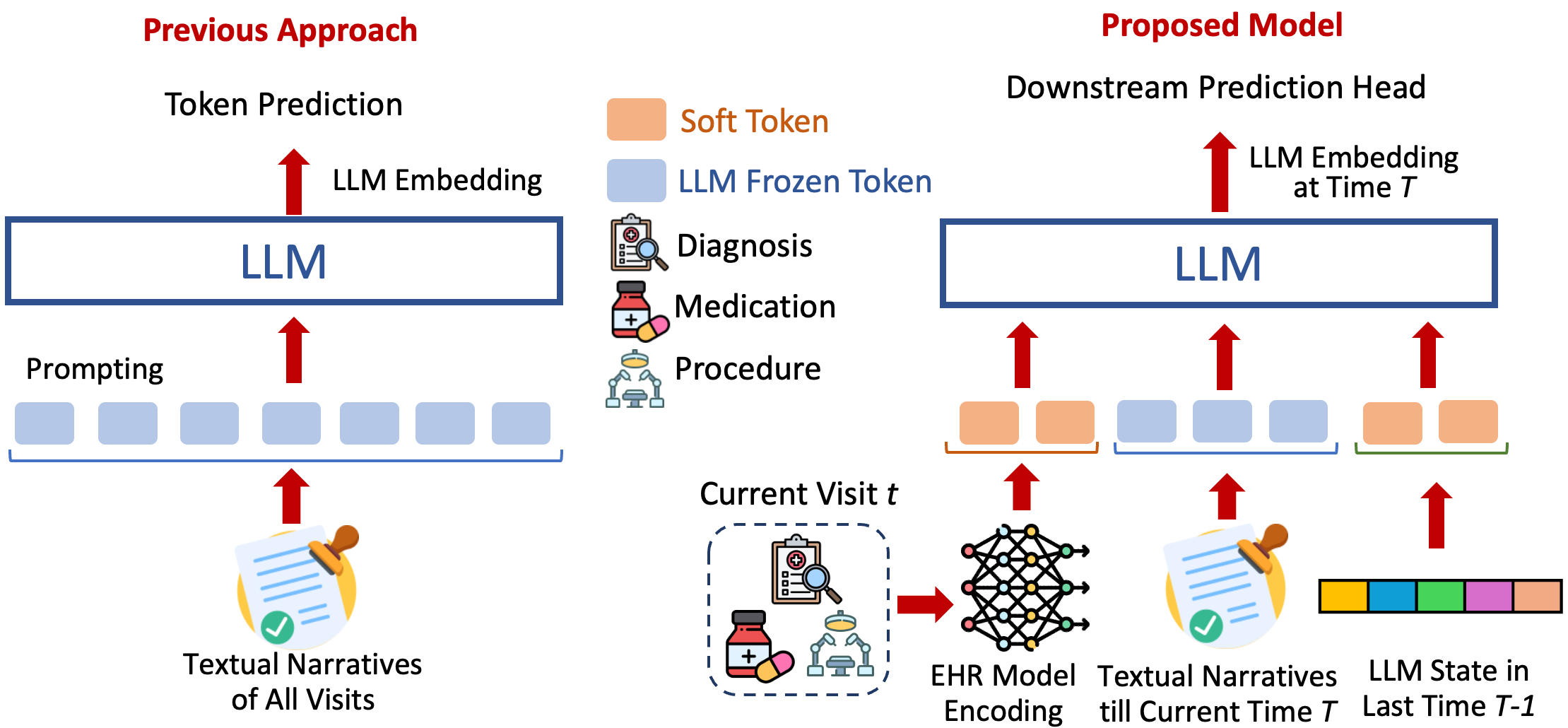}
	\caption{Illustration of the difference between existing approaches based on hard prompting and the proposed RePrompT framework. Unlike prior methods, RePrompT integrates both hard and soft prompting, with the soft prompts implemented through two strategies: struct-encoder and state-recurrent prompting.
    }
    \label{llmVSLLMClin}
    \vspace{-0.2cm}
\end{figure}


\begin{itemize}[leftmargin=*]

\item We develop a recurrent prompt tuning mechanism that allows an LLM to propagate visit-level EHR representation by reusing hidden states across time steps. This design mitigates the limitation of standard LLM inference, where the visit structure is only weakly encoded in plain text.

\item We propose a framework that integrates general-purpose LLMs with cohort-trained, task-aligned EHR encoders by injecting structured EHR embeddings as trainable prompt tokens. By grounding the LLM in a shared patient representation space, our method enables population-aware modeling, in contrast to existing inference approaches that rely solely on text linearization.

\item We conduct extensive experiments on two large-scale public benchmarks, MIMIC-III and MIMIC-IV, across readmission and mortality prediction tasks. Results show that the proposed approach consistently outperforms strong EHR-based and LLM-based baselines across different tasks and datasets.



 
\end{itemize}

\section{Methodology}
\begin{figure*}[ht]
        \centering
	\includegraphics[width= 0.95\textwidth]{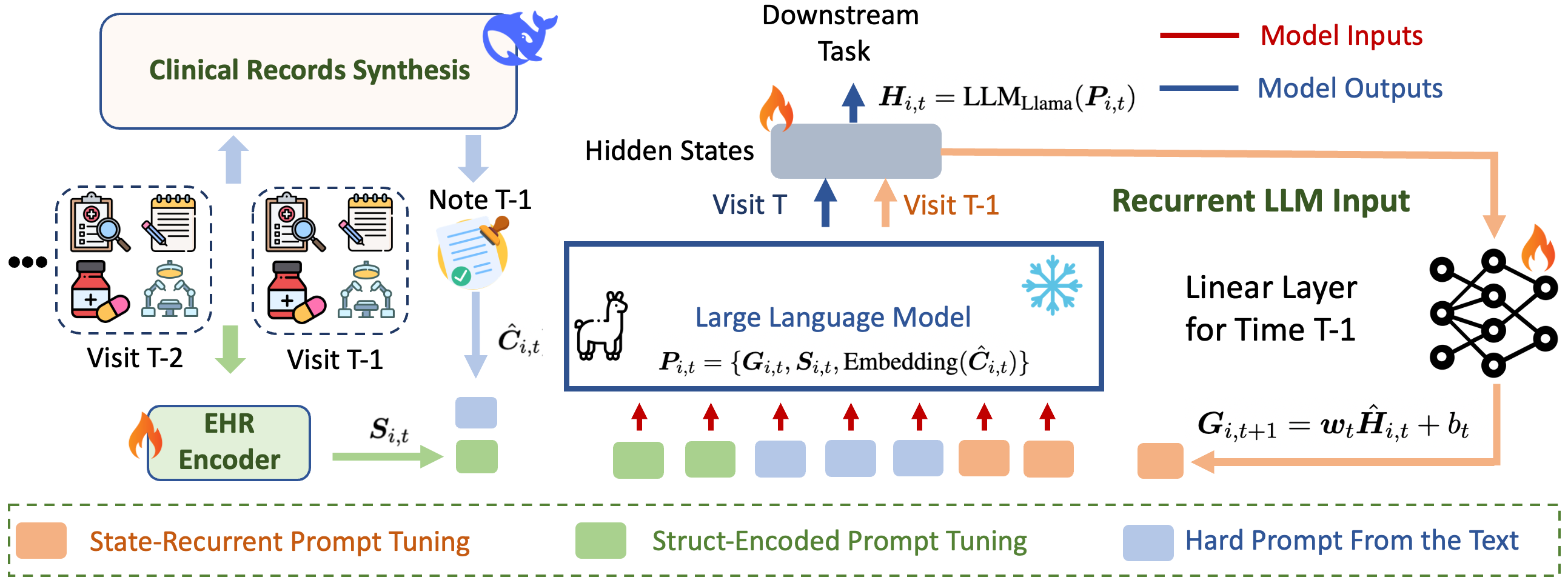}
    \caption{The framework of the proposed method. Medical codes and discharge notes are first used to generate patient summaries through Clinical Records Synthesis. Next, the structured medical codes in patient history are encoded into embeddings through a classic EHR encoder. Meanwhile, the LLM’s hidden state in the previous time step is recurrently taken as a soft prompt for the current time step to capture longitudinal dependencies. 
    Outputs from all prompting methods are combined as the input of the predictive LLM, and the LLM’s state at the final visit is used for downstream binary classification tasks.}
    \label{method}
    \vspace{-0.2cm}
\end{figure*}

\subsection{Problem Formulation}
The EHR data for patient $i$ is represented as a sequence of clinical visits 
$\{V_{i,t}\}_{t=1}^{T_i}$, where $V_{i,t}$ denotes the $t$-th visit in chronological order, 
and $T_i$ is the total number of visits for patient $i$. Each visit $V_{i,t}$ consists of a 
set of medical codes, including diagnoses, medications, and procedures. Formally, the set of 
codes for visit $t$ is defined as $V_{i,t} = \{x^{i}_{j,t}\}_{j=1}^{|V_{i,t}|}$, where $|V_{i,t}|$ denotes the number of medical codes recorded during that visit. In addition to medical codes, each visit also contains discharge notes represented as textual summaries. We denote the set of tokens in the discharge note for patient $i$ at visit $t$ as
$\textbf{C}_{i,t} = \{c^{i}_{j,t}\}_{j=1}^{|C_{i,t}|}$, where each $c^{i}_{j,t}$ corresponds to a discrete token. In this research, the terms ``time-steps'' and ``visits'' refer to the same concepts.

\noindent\textbf{Task:} 
Given a patient $i$ with a sequence of visits, where each visit contains 
a set of medical codes $\{V_{i,t}\}_{t=1}^{T_i}$, and a corresponding discharge note $\{C_{i,t}\}_{t=1}^{T_i}$, the objective is to predict a specific healthcare outcome (e.g., mortality, readmission, or medication) in 
the next visit at $T_i+1$. This prediction is formulated as a binary or multi-label classification task for the target
$Y_{i, T_i+1}$.
\subsection{Model Summary}
As shown in Figure \ref{method}, the proposed approach consists of three main modules: (1) Clinical Records Synthesis: Given a patient’s medications, procedures, diagnosis codes, and discharge notes, we first prompt a powerful and general-purpose LLM (e.g., DeepSeek) to synthesize a comprehensive patient summary. (2) State-Recurrent Prompt Tuning: We then use a local tunable LLM (e.g., Llama) to generate the hidden state from the previous time step and propagate it as a soft prompt to guide the generation at the current time step to capture longitudinal dependencies across visits. (3) Struct-Encoded Prompt Tuning: In parallel, we adopt a structured EHR encoder \cite{retain} to encode the sequential history of medical codes into dense representations, which serve as another set of soft prompts, allowing the tunable LLMs to incorporate shared structured patterns across different patient cases.
Conditioned on both the synthesized patient summary and the two complementary soft prompting strategies, the proposed model generates a final representation for downstream classification tasks. 

\subsection{Clinical Record Synthesis}
Structured EHRs consist of standardized medical codes, including medications, diagnoses, and procedures across patient visits. In addition, each visit (e.g., hospital stays) is accompanied by a discharge note that summarizes the
patient’s clinical course \cite{MIMIC3, johnson2023mimic}. These notes are often verbose and noisy, containing redundant information and templated sections, which limit their usefulness for downstream modeling.

\begin{figure}[t]
        \centering
	\includegraphics[width= 8cm]{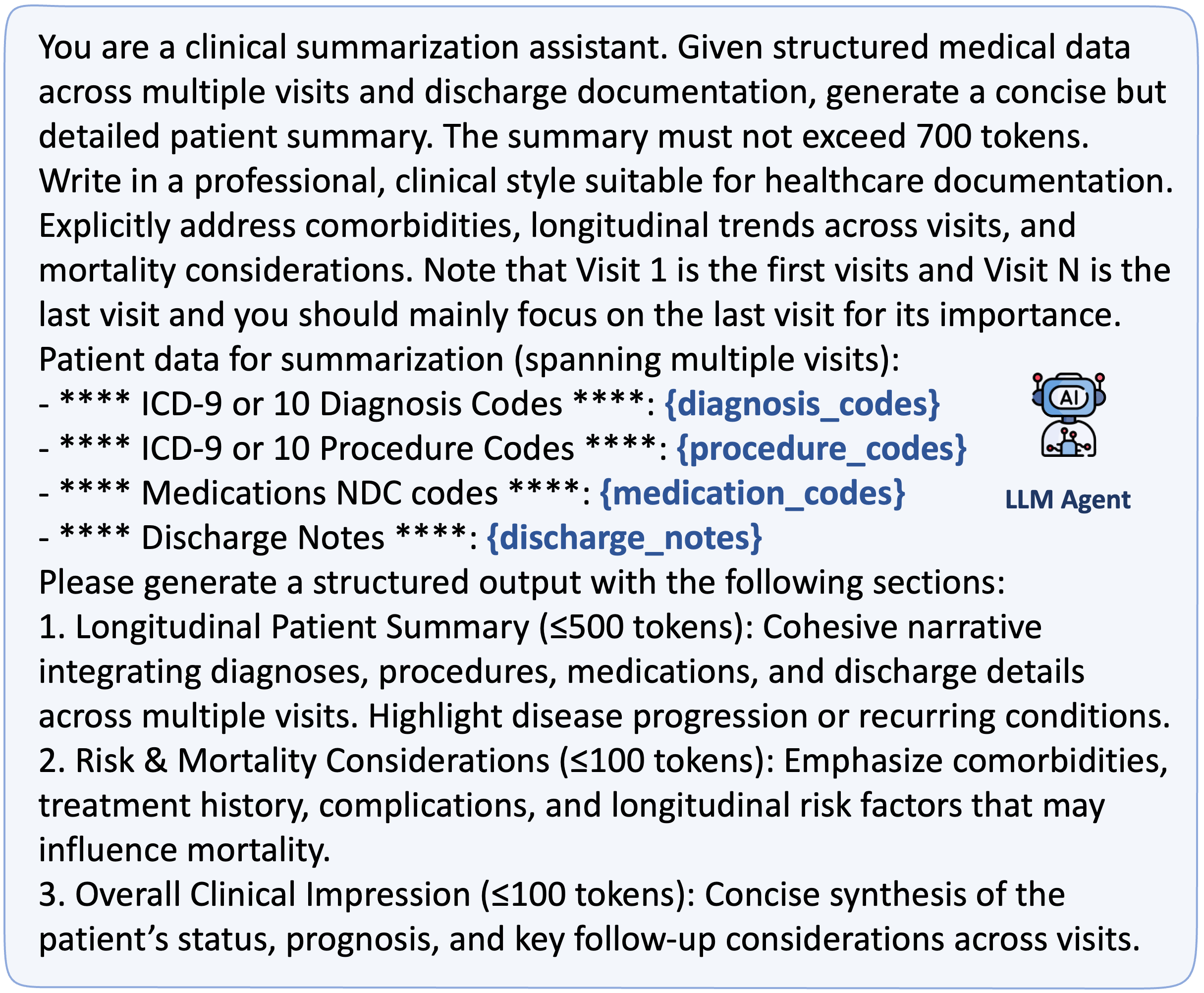}
	\caption{The Prompt for clinical record synthesis.}
    \label{prompt}
    \vspace{-0.5cm}
\end{figure}

To address this, we employ a general-purpose LLM to denoise and synthesize discharge notes together with structured medical codes into a concise patient summary. Specifically, we use DeepSeek-V3 \cite{liu2024deepseek} to summarize information from historical clinical notes and corresponding visit-level codes, as illustrated in Figure \ref{prompt}. The same synthesis prompt is applied at each visit $t$ to produce a unified patient representation:
\begin{equation} 
\label{eq:llm_transform} 
\boldsymbol{\hat{C}_{i, t}} = \text{LLM}_{\text{DeepSeek}} ( \{ \boldsymbol{V}_{i,r}, \boldsymbol{C}_{i,r} | r <= t\}),
\end{equation} 
where $\boldsymbol\hat{C}_{i,t}$ denotes the textual narratives of patient $i$ at time step $t$.

\subsection{Prompt Tuning for LLMs}
In a standard language model, discrete input tokens are first mapped into continuous vectors through an embedding layer, which are then processed by the subsequent self-attention layers to produce the model outputs. 
Achieving strong performance on a specific downstream task often requires adapting these models through fine-tuning on task datasets.
Beyond conventional approaches such as full fine-tuning or parameter-efficient methods like LoRA \cite{hu2022lora}, prompt tuning \cite{lester2021power} offers an effective alternative.
Rather than updating the LLM's internal weights, prompt tuning focuses on generating a small set of trainable continuous vectors, often called soft prompts, as the input embedding sequence to steer the model toward the target task. Because only the prompt parameters are optimized, prompt tuning reduces the cost and complexity of adaptation and is particularly appealing when the base LLM is large or not fully accessible, while still retaining strong flexibility across tasks.

Another practical advantage of soft prompts lies in their flexibility: since they are represented as trainable continuous variables, this method helps LLMs to be seamlessly integrated with existing neural architectures, allowing modular extensions without altering the original LLM parameters.
Formally, the LLM input at time step $t$ is:  
\begin{equation} 
\label{eq:llm_model} 
\boldsymbol{P}_{i,t} = \{ \boldsymbol{G}_{i,t}, \boldsymbol{S}_{i,t}, \text{Embedding}(\hat{\boldsymbol{C}}_{i,t}) \}, 
\end{equation}  
where $\boldsymbol{G}_{i,t} \in \mathcal{R}^{P \times D}$ corresponds to the \textit{State-Recurrent Prompt Tuning} module, $D$ is the hidden dimension of the LLM, and $P$ represents the number of soft prompts for each module. $\boldsymbol{S}_{i,t} \in \mathcal{R}^{P \times D}$ is derived from the \textit{Struct-Encoded Prompt Tuning} module, and $\text{Embedding}(\hat{\boldsymbol{C}}_{i,t}) \in \mathcal{R}^{N_{i, t} \times D}$ represents the token embedding associated with the \textit{Clinical Record Synthesis}, where $N_{i, t}$ is the number of tokens for the patient summary for patient $i$ at visit $t$. As shown in Equation \ref{eq:llm_embed_downstream}, the representation $\boldsymbol{P}_{i,t}$ is subsequently fed into a predictor LLM, which produces the hidden state $H_{i,t} \in \mathbb{R}^{(2P + N_{i,t}) \times D}$. This hidden state is then used for the downstream classification.
\begin{equation} 
\label{eq:llm_embed_downstream} 
\boldsymbol{H}_{i,t} = \text{LLM}_{\text{Llama}}(\boldsymbol{P}_{i,t})
\end{equation}  

Since our primary focus is on embedding generation, we employ the Llama 3.1 1B model using the LLM2Vec framework, which adapts the language model specifically for high-quality embedding extraction \cite{behnamghader2024llm2vec}. 

\subsection{State-Recurrent Prompt Tuning}
LLMs have proven effective at converting textual information or prompts into high-level embedding representations, which can then be utilized to generate the next token. In existing literature, the input to an LLM is typically treated as a single document that is processed through multiple layers of self-attention, enabling the model to produce responses token by token. While powerful, this design introduces limitations in the context of EHR.

In EHR data, each patient has multiple visits, and within each visit, there is often a corresponding discharge note written by the physician. A straightforward approach to formatting sequential data for LLMs is to concatenate the visits into raw text with markers such as “Visit 1: … Visit 2: …”. Although this method distinguishes visits through text description, the LLM still inherently treats the input as a single document and lacks an explicit mechanism to associate tokens with specific temporal dependencies. 

To address this limitation, we propose State-Recurrent Prompt Tuning, an approach designed to better capture the structure of the visit level and longitudinal patient trajectories within EHR. Instead of aggregating all visit information as a single input to the LLM, we make the LLM process only one visit at a time, and output the token-level hidden state $H_{i,t}$ from the last layer, which is then aggregated through average pooling to form a visit-level hidden state $\hat{H}_{i,t}$.
This state vector serves as a soft prompt that will be recurrently passed back to the same LLM to guide the generation of the hidden state for the subsequent visit, thereby enabling temporal continuity across visits.

As formulated in Equation \ref{eq:linear}, we first apply a linear transformation to the pooled hidden state, where $\boldsymbol{w}_t$ and $b_t$ are trainable parameters, and $\boldsymbol{G}_{i,t} \in \mathbb{R}^{P \times D}$ shows the soft prompt embeddings, with $P$ denoting the number of soft prompts.
\begin{equation} 
\label{eq:linear} 
\boldsymbol{G}_{i, t+1} = \boldsymbol{w}_t \hat{\boldsymbol{H}}_{i, t} + b_t
\end{equation} 

The output of this module, together with the following prompt tuning component will construct the soft prompt of the LLM.

\subsection{Struct-Encoded Prompt Tuning}

%

LLM-based prompting methods for EHR mining typically linearize extensive patient histories spanning multiple visits into a single, exhaustive per-patient input for LLMs to process \cite{healthllm3, healthllm7}. While this formulation allows LLMs to access rich clinical narratives, it introduces two fundamental limitations. First, collapsing longitudinal records into static text hinders the model’s ability to capture evolving patient trajectories and disease progression over time, as temporal dependencies across visits are not explicitly represented \cite{healthllm7}. Second, such per-patient text representations prevent LLMs from effectively leveraging population-level and task-specific patterns that are critical for clinical prediction. Unlike cohort-trained EHR models, LLMs optimized with token- or sequence-level objectives do not enforce patient-centric alignment across the cohort, making it difficult to encode disease co-occurrence, shared ontologies, and longitudinal similarities among patients \cite{healthllm3}. Consequently, clinically similar patients are not guaranteed to occupy nearby regions in the representation space, which leads to less contextually rich embeddings. Naively incorporating additional patient histories into the prompt is infeasible due to window constraints.

Therefore, we leverage structured encoders that learn patient representations from sequences of medical codes. Such models compress a patient’s longitudinal history into a dense embedding that captures clinically meaningful patterns shared across patients. This representation can be injected into the LLM as a soft prompt. Among existing approaches, we adopt RETAIN \cite{retain} due to its effective use of dual-level attention and recurrent modeling for summarizing patient histories. Architectural details are provided in Appendix~\ref{app:retain}. Given visit embeddings $\{\boldsymbol{V}_{i,j}\}_{j=1}^{t}$ RETAIN produces a patient representation $\boldsymbol{S}_{i,t}$, which is used as input to the LLM.

\subsection{Prediction and Optimization}
The output layer of an LLM is designed for next-token prediction. However, it can not reliably provide the calibrated probabilities needed for the classification tasks, as probabilities are treated as tokens and often suffer from hallucinations \cite{wang2024calibrating}. To address this limitation, we introduce a classification head on top of the LLM, which takes the hidden state corresponding to the last visit input and maps it directly to one or multiple binary classes. Formally, as expressed in Equation \ref{eq:output}, the model produces the output $\mathbf{y}_{i, T_{i+1}}$, which is then passed through the linear layer to generate scores. A threshold is then applied to yield the prediction $Y_{i, T_{i+1}}$.
\begin{equation} 
\label{eq:output} 
\mathbf{y}_{i, T_{i+1}} = \boldsymbol{w}_{\text{output}} \hat{\boldsymbol{H}}_{i, T_i} + b_{\text{output}}
\end{equation}

For optimization, we employ the Adam optimizer \cite{loshchilov2017decoupled}, and for the loss function, we use Binary Cross-Entropy loss in binary and multi-label classification tasks.

The trainable components include the EHR encoder, implemented using the RETAIN model (Equations 6-11), the linear layer that transforms the hidden representation at time $T_{i}-1$ into the input representation at time $T_{i}$
 (Equation 4), and the output layer that maps the LLM output representation to the classification head (Equation 5). The LLaMA model remains frozen during both training and inference.
\section{Experimental Setup}

\subsection{Datasets}
In this study, we employ two real-world datasets to evaluate both mortality prediction and readmission prediction tasks:

\begin{table}[t]
\centering
\caption{Statistics of the EHR datasets.}
\label{Statistical}
\small
\renewcommand{\arraystretch}{1.1}
\setlength{\tabcolsep}{3pt} 
\makebox[\linewidth]{
    \begin{tabular}{@{}lcc@{}}
    \hline
    Metric & \multicolumn{1}{c}{\bf{MIMIC-III}} & \multicolumn{1}{c}{\bf{MIMIC-IV}}\\
    \hline
    \# of patients & 7537 & 15874  \\
    \# of patients with 2 visits & 3622 & 4991  \\
    \# of drugs per patient  & 79.30 & 113.55  \\
    \# of diagnosis per patient & 28.98  & 65.99  \\
    \# of procedures per patient & 7.37 & 9.08  \\
    \# of visits per patient   & 1.65 & 3.11  \\
    \# of patients with 1 visit & 3622 & 4991\\
    Positive rate for readmission &  53.7\% &  53.5\% \\
    Positive rate for mortality & 6.6\% & 1.3\% \\
    \hline
    \end{tabular}
}
\vspace{-0.2cm}
\end{table}

\noindent $\bullet$ \textbf{MIMIC-III} \cite{MIMIC3} is an open-access database containing health records of over 40,000 patients admitted to critical care units between 2001 and 2012. In this study, we focus on patients with multiple visits, aiming to predict the binary outcome.\\
\noindent $\bullet$ \textbf{MIMIC-IV} \cite{johnson2020mimic} is a publicly available EHR dataset covering hospital admissions at Beth Israel Deaconess Medical Center from 2008 to 2019. It extends MIMIC-III with a clearer modular structure, richer clinical detail, and improved data provenance. MIMIC-IV contains data on over 380,000 unique patients across.

\begin{table*}[ht]
\centering
\renewcommand{\arraystretch}{1.1}
\footnotesize
\setlength{\tabcolsep}{3pt}
\begin{tabular}{l|cc|cc||cc|cc}
\hline
\multirow{2}{*}{\textbf{Model}}
& \multicolumn{4}{c||}{\textbf{MIMIC-IV}}
& \multicolumn{4}{c}{\textbf{MIMIC-III}} \\ \cline{2-9}
& \multicolumn{2}{c|}{\textbf{Readmission}}
& \multicolumn{2}{c||}{\textbf{Mortality}}
& \multicolumn{2}{c|}{\textbf{Readmission}}
& \multicolumn{2}{c}{\textbf{Mortality}} \\ \cline{2-9}
& AUROC & PRAUC & AUROC & PRAUC & AUROC & PRAUC & AUROC & PRAUC \\ \hline

Deepr
& 0.614 & 0.647 & \uline{0.667} & 0.028
& 0.673 & \uline{0.705} & \uline{0.638} & 0.140 \\

RETAIN
& \uline{0.670} & 0.690 & 0.601 & 0.031
& 0.660 & 0.676 & 0.608 & 0.134 \\

GRAM
& 0.578 & 0.609 & 0.633 & 0.028
& 0.627 & 0.660 & 0.626 & 0.139 \\

GRASP
& 0.537 & 0.572 & 0.666 & 0.029
& 0.617 & 0.639 & 0.632 & 0.137 \\

AdaCare
& 0.608 & 0.641 & 0.643 & 0.029
& 0.640 & 0.681 & 0.609 & 0.141 \\

StageNet
& 0.656 & 0.691 & 0.664 & 0.032
& \uline{0.676} & 0.702 & 0.633 & \uline{0.142} \\

Adore
& 0.605 & 0.641 & 0.656 & 0.024
& 0.661 & 0.693 & 0.629 & 0.138 \\

ARCI
& 0.663 & \uline{0.692} & 0.611 & \uline{0.034}
& 0.652 & 0.671 & 0.618 & 0.129 \\

\textbf{RePrompT}
& \textbf{0.706} & \textbf{0.728} & \textbf{0.673} & \textbf{0.036}
& \textbf{0.688} & \textbf{0.719} & \textbf{0.646} & \textbf{0.152} \\

\hline
\end{tabular}
\caption{Performance comparison on readmission and mortality prediction tasks using the MIMIC-III and MIMIC-IV datasets. Evaluation is conducted based on AUROC and PRAUC metrics.}
\label{tab:merged_results}
\end{table*}

Both datasets are publicly available and have been thoroughly de-identified to comply with U.S. HIPAA regulations, which mandate the removal or modification of 18 types of personal identifiers. Their use in this study was conducted under the PhysioNet credentialed data use agreement.

\subsection{Implementation Details}
In this study, we focus on predicting two key healthcare outcomes: hospital readmission and mortality. Specifically, given information up to visit $T_i$, the model predicts the binary outcome at visit $T_{i+1}$ for patient $i$ with $T_i$ prior visits or recommends medications on visit $T_i$ based on diagnosis and procedures available from time $1$ to $T_i$. 
Since these task requires longitudinal data, we excluded patients with only a single recorded visit.

For RePrompT and baselines, we use a greedy search approach to find the best hyperparameters for a comprehensive evaluation. We randomly split the data into 70\% training and 30\% testing sets and report the mean over three runs. We found that the optimal number of soft prompts is $P = 10$ or both modules, based on experiments with varying numbers of soft prompts for each component, which showed that this setting provides a favorable balance between performance and complexity. We also set the hidden dimension for the EHR model to 256 for the RETAIN model.

The dataset statistics are presented in Table \ref{Statistical}. All experiments are implemented in Python, with PyTorch \cite{pytorch} serving as the primary deep learning framework. In addition, RePrompT is fully compatible with the PyHealth framework \cite{pyhealth}, from which we use EHR baselines implementations. For LLM tuning we use the Hugging Face \cite{wolf2019huggingface} framework. We utilize a high-performance server equipped with three NVIDIA A6000 GPUs, 256 GB of RAM, and a 48-core CPU. We release the source code.\footnote{\url{https://github.com/KU-AI4H/RePrompT}} The computation time for a batch of patients is detailed in Appendix \ref{app:comp_ta}.

In this research, we used two well-known threshold-independent classification metrics to comprehensively evaluate RePrompT, namely the AUROC and the PRAUC scores for readmission and mortality prediction.

\begin{table*}[ht]
\centering
\small
\renewcommand{\arraystretch}{1.2}
\setlength{\tabcolsep}{4pt}
\begin{tabular}{l|cc|cc||cc|cc}
\hline
\multirow{3}{*}{\textbf{Model}} 
& \multicolumn{4}{c||}{\textbf{MIMIC-IV}} & \multicolumn{4}{c}{\textbf{MIMIC-III}} \\ \cline{2-9}
& \multicolumn{2}{c|}{\textbf{Readmission}} & \multicolumn{2}{c||}{\textbf{Mortality}} 
& \multicolumn{2}{c|}{\textbf{Readmission}} & \multicolumn{2}{c}{\textbf{Mortality}} \\ \cline{2-9}
& AUROC & PRAUC & AUROC & PRAUC & AUROC & PRAUC & AUROC & PRAUC \\ \hline
Zero-Shot  & 0.512 & 0.545 & 0.591 & 0.011 & 0.505 & 0.537 & 0.501 & 0.069  \\
Prompt-Tuning      & 0.575 & 0.608 & \uline{0.607} & 0.019 & 0.634 & 0.663 & 0.611 & 0.132 \\
COCONUT      & \uline{0.581} & \uline{0.611} & 0.605 & \uline{0.021} & \uline{0.639} & \uline{0.670} & \uline{0.612} & \uline{0.136} \\
\textbf{RePrompT}      & \textbf{0.706} & \textbf{0.728} & \textbf{0.673} & \textbf{0.036} & \textbf{0.688} & \textbf{0.719} & \textbf{0.646} & \textbf{0.152} \\
\hline
\end{tabular}
\caption{Performance comparison of LLM-based baselines on the MIMIC-III and MIMIC-IV datasets for both readmission and mortality prediction tasks on PRAUC and AUROC.}
\label{tab:llm_baselines}
\end{table*}

\subsection{Baselines}
\vspace{-0.2cm}
In this research, we utilize both well-known healthcare deep learning methods and LLM-based approaches. For deep learning methods, we use the following models  (1) \noindent \textbf{Deepr} \cite{nguyen2016mathtt} represents each patient record as a sequence of coded events with time gaps and hospital transfers, then applies a convolutional neural network (2) \noindent \textbf{RETAIN} \cite{retain} incorporates a dual-RNN network to capture the interpretable influence of the visits and medical features for the prediction tasks. (3) \noindent \textbf{GRAM} \cite{GRAM} is a graph-based attention model that enriches EHR data with the hierarchical medical ontology, representing each concept as a weighted combination of its ancestors. (4) \noindent \textbf{GRASP} \cite{zhang2021grasp} is a healthcare framework that improves EMR-based prediction by finding clinically similar patients. (5) \noindent \textbf{AdaCare} \cite{ma2020adacare} is a health status representation model that captures both short- and long-term biomarker variations, adaptively emphasizes patient-specific risk factors. (6) \noindent \textbf{StageNet} \cite{ma2020adacare} is a stage-aware neural network that learns disease progression via LSTM and integrates them with stage-adaptive convolution. (7) \noindent \textbf{ADORE} \cite{cheong2023adaptive}  uses attention to adapt medical ontology category embeddings to EHR data for improved clinical prediction. (8) \noindent \textbf{ARCI} \cite{ARCI}  disentangles coexisting temporal medical intents across sequential visits.

We have also conducted experiments on three different LLM-based baselines: (1) \textbf{Zero Shot} \cite{zhu2024prompting}, where GPT-5 \cite{wang2025capabilities} is prompted to output probabilities for mortality and readmission prediction. (2) \textbf{Prompt-Tuning}, which introduces trainable soft prompts without relying on an EHR encoder and gets the probability of the “Yes” token from next-token prediction as the model output using the Llama 3.1 1B model. \cite{lester2021power} (3) \textbf{COCONUT} \cite{hao2024training}, where a soft token is generated prior to predicting “Yes” or “No” to incorporate temporal information with the Llama 3.1 1B model, similar to our proposed approach.

\section{Results and Discussion}
This section presents the experimental analysis of comparisons between the proposed method and EHR and LLM-based approaches, ablation studies on the model and summarization, and evaluations using different EHR encoders. 



\subsection{Performance Comparison with EHR Baselines}
Table \ref{tab:merged_results} presents a comprehensive comparison between the proposed RePrompT framework and several well-established EHR baselines on both the MIMIC-III and MIMIC-IV datasets across two binary classification tasks.
The results consistently highlight the superior performance of RePrompT across all evaluation metrics. In particular, when compared to RETAIN on the mortality prediction task, RePrompT achieves a substantial performance gain. This improvement stems from the use of a time-aware prompt tuning strategy that effectively links patient-specific EHR embeddings to the LLM to have more accurate modeling of longitudinal patient trajectories. Furthermore, against StageNet, the strongest baseline for MIMIC-IV mortality prediction, RePrompT demonstrates clear advantages. By integrating attention-aware RNNs with LLMs, our method surpasses the hybrid RNN-CNN architecture of StageNet, underscoring the benefit of incorporating language models into temporal EHR representations. Finally, the comparison with GRASP reveals that the time-aware LLM approach captures richer and more clinically meaningful information than methods relying solely on patient similarity during the embedding generation phase. 


\begin{table*}[t]

\centering

\resizebox{\textwidth}{!}{
\begin{tabular}{l|cc|cc||cc|cc}
\hline
\multirow{3}{*}{\textbf{Model}} 
& \multicolumn{4}{c||}{\textbf{MIMIC-IV}} 
& \multicolumn{4}{c}{\textbf{MIMIC-III}} \\
\cline{2-9}
& \multicolumn{2}{c|}{\textbf{Readmission}} 
& \multicolumn{2}{c||}{\textbf{Mortality}} 
& \multicolumn{2}{c|}{\textbf{Readmission}} 
& \multicolumn{2}{c}{\textbf{Mortality}} \\
\cline{2-9}
& AUROC & PRAUC & AUROC & PRAUC & AUROC & PRAUC & AUROC & PRAUC \\
\hline
RePrompT w/o Both Modules 
& 0.673 & 0.698 & 0.635 & 0.028 & 0.646 & 0.692 & 0.616 & 0.127 \\

RePrompT w/o State-Recurrent 
& 0.693 & 0.711 & 0.642 & \uline{0.033} & 0.673 & 0.705 & 0.624 & 0.132 \\

RePrompT w/o Struct-Encoded 
& \uline{0.698} & \uline{0.722} & \uline{0.665} & 0.031 & \uline{0.676} & \uline{0.712} & \uline{0.637} & \uline{0.148} \\

\textbf{RePrompT} 
& \textbf{0.706} & \textbf{0.728} & \textbf{0.673} & \textbf{0.036} 
& \textbf{0.688} & \textbf{0.719} & \textbf{0.646} & \textbf{0.152} \\
\hline
\end{tabular}
}
\caption{Ablation Studies of the proposed method RePrompT on the MIMIC-III and MIMIC-IV datasets for both readmission and mortality prediction tasks on PRAUC and AUROC. }
\vspace{-0.1cm}
\label{tab:RePrompT_ablation}
\end{table*}

\begin{figure}[t]
        \centering
	\includegraphics[width= 7.5cm]{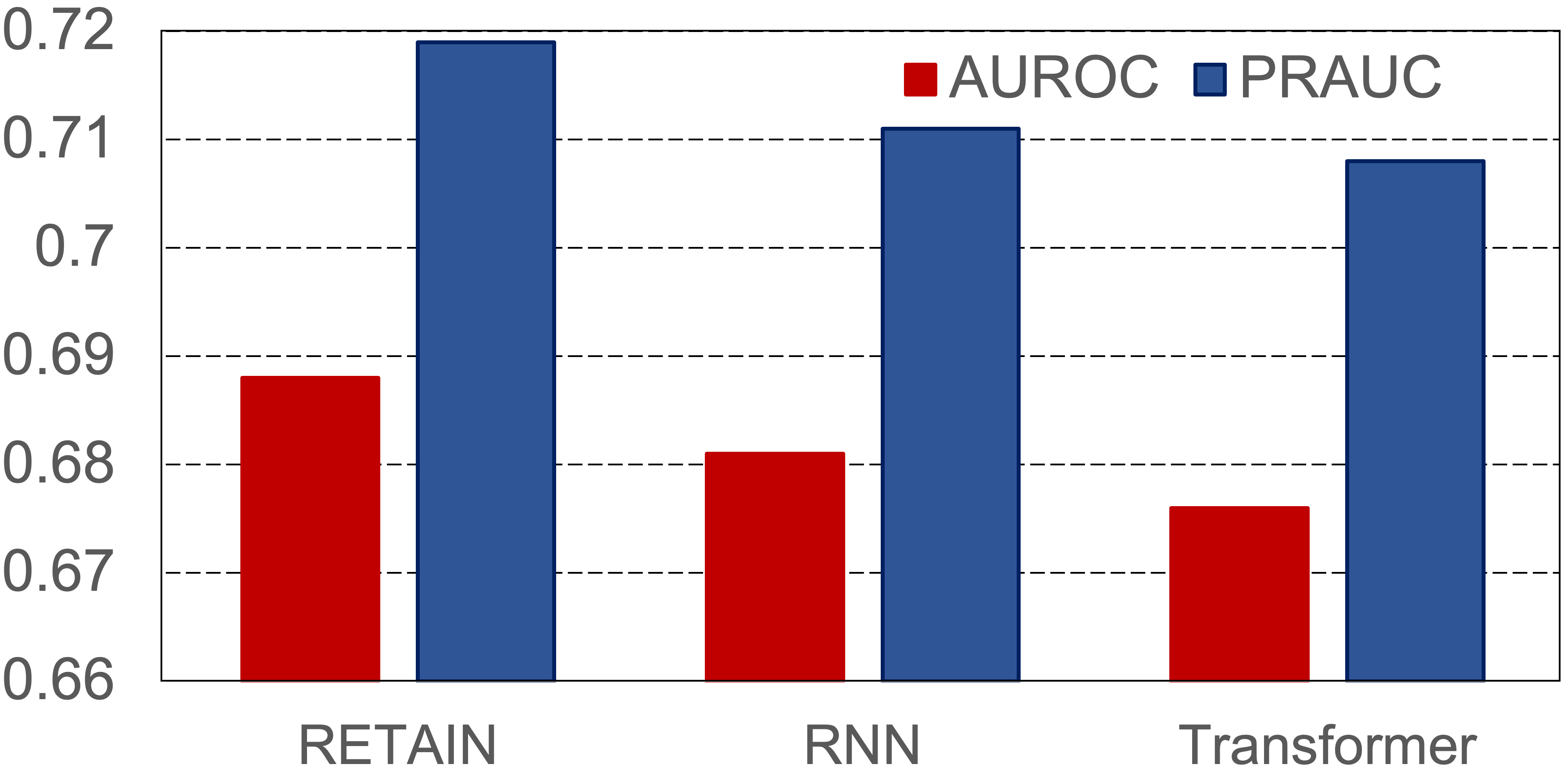}
    \caption{Performance comparison of different EHR encoders integrated into RePrompT. Results are reported using PRAUC and AUROC for the task of readmission prediction on the MIMIC-III dataset.}
    \label{fig:both_analysis}
\end{figure}

\subsection{Performance Comparison for Different Integrated EHR Models}
As illustrated in Figure \ref{fig:both_analysis}, we conducted a series of complementary experiments focusing on the EHR backbone integrated within RePrompT for readmission and mortality prediction in the MIMIC-III dataset. In this analysis, patient embeddings generated by the EHR model were extracted and subsequently utilized as soft prompts within the LLM component, providing a more fine-grained examination of how different EHR architectures influence overall performance. Among the evaluated models, RETAIN achieved the highest performance, highlighting the effectiveness of its attention mechanism in emphasizing clinically relevant information from recent visits. For comparison, we also assessed two alternative configurations: one employing a standard LSTM architecture and another using a Transformer Encoder \cite{vaswani2017attention} as the EHR module. The Transformer-based approach performed the worst, likely because Transformer encoders have difficulty in modeling temporal dependencies across successive patient visits, even when positional encodings are applied \cite{zhou2021informer}. 

\subsection{Performance Comparison with LLM Baselines}
Table \ref{tab:llm_baselines} presents a comparison between the proposed approach and several well-established LLM-based baselines across both datasets. When the proposed approach is contrasted with the Zero-Shot Prompt Engineering method, our time-aware prompt tuning strategy achieves substantially higher performance on EHR prediction tasks, confirming that fine-tuned smaller models outperform larger models in specific tasks \cite{gao2023small, bucher2024fine}.  Furthermore, comparisons with prompt-tuning baselines show the critical importance of integrating soft prompts with the EHR model. Standard LLMs lack explicit knowledge of medical code co-occurrence patterns and thus fail to fully capture clinically meaningful embeddings. While chain-of-thought reasoning methods, such as the COCONUT approach, which also employs soft prompts, are generally advantageous, their performance here lags behind because conventional chain-of-thought reasoning does not adequately model temporal dependencies. In contrast, our time-aware chain-of-thought variant more effectively captures the evolving nature of patient trajectories, leading to superior performance. 

\subsection{Ablation Studies on RePrompT}
Table \ref{tab:RePrompT_ablation} reports the ablation results used to assess the impact of each component in our framework. Removing either module leads to consistent performance degradation across all datasets and tasks, indicating that both components contribute meaningfully to the final model. The State-Recurrent Network produces the larger performance drop when excluded, suggesting that explicit modeling of temporal dependencies at the textual level plays a central role in improving predictive accuracy. Meanwhile, the gains associated with the Struct-Encoded module highlight the benefit of incorporating structured time-series information from multi-level EHR data. Taken together, these results show that the two modules capture complementary signals and jointly strengthen the model’s ability to use multimodal clinical information.

\begin{table*}[ht]
\centering
\small
\renewcommand{\arraystretch}{1.2}
\setlength{\tabcolsep}{4pt}
\begin{tabular}{l|cc|cc||cc|cc}
\hline
\multirow{3}{*}{\textbf{Model}} 
& \multicolumn{4}{c||}{\textbf{MIMIC-IV}} & \multicolumn{4}{c}{\textbf{MIMIC-III}} \\ \cline{2-9}
& \multicolumn{2}{c|}{\textbf{Readmission}} & \multicolumn{2}{c||}{\textbf{Mortality}} 
& \multicolumn{2}{c|}{\textbf{Readmission}} & \multicolumn{2}{c}{\textbf{Mortality}} \\ \cline{2-9}
& \textbf{AUROC} & \textbf{PRAUC} & \textbf{AUROC} & \textbf{PRAUC} 
& \textbf{AUROC} & \textbf{PRAUC} & \textbf{AUROC} & \textbf{PRAUC} \\ \hline

RETAIN & 0.670 & 0.690 & 0.601 & 0.031 & 0.660 & 0.676 & 0.608 & 0.134 \\
RePrompT (w/o DeepSeek) & 0.685 & 0.705 & 0.640 & 0.027 & 0.670 & 0.702 & 0.631 & 0.135 \\
RePrompT & \textbf{0.706} & \textbf{0.728} & \textbf{0.673} & \textbf{0.036} & \textbf{0.688} & \textbf{0.719} & \textbf{0.646} & \textbf{0.152} \\

\hline
\end{tabular}
\caption{Ablation Studies of input summerization on the MIMIC-III and MIMIC-IV datasets for both readmission and mortality prediction tasks on PRAUC and AUROC.}
\label{tab:DSeek}
\end{table*}

\subsection{Ablation Study on Input Summarization}
\label{app:summerize}
To verify whether the improvement is mainly due to the DeepSeek-generated hard prompt, we removed DeepSeek and used only clinical notes as input to the Llama model. As shown in Table \ref{tab:DSeek}, the proposed method without the DeepSeek summarization still outperforms the RETAIN backbone, showing that the improvement comes from the framework itself, not only from DeepSeek summerization. The results also suggest that denoising the notes can improve performance.


\section{Related Work}
\textbf{EHR-Based Predictive Models.}
Previous works have explored transforming EHRs into predictive representations for clinical decision support. Early deep learning models such as Deepr \cite{nguyen2016mathtt} bypass manual feature engineering by encoding records as sequences of discrete events, with convolutional networks detecting predictive clinical motifs for readmission risk. RETAIN \cite{retain} enhances interpretability with a reverse-time attention mechanism that highlights influential visits and variables, mimicking how clinicians review patient histories. To address data sparsity and domain alignment, subsequent methods incorporate external knowledge. GRAM \cite{GRAM} leverages hierarchical ontologies to produce knowledge-aligned embeddings, improving prediction for rare conditions. Hierarchical Attention Propagation (HAP) \cite{hap} extends this idea by propagating attention bidirectionally across the ontology, capturing relationships among ancestors and descendants. While effective, such approaches do not naturally capture temporal dependencies across visits. GRASP \cite{zhang2021grasp} embeds medical concepts into a unified semantic space using large language models, aligning semantically similar codes across datasets and mitigating coding inconsistencies. Patient heterogeneity and disease progression have been addressed by AdaCare \cite{ma2020adacare}, which models short- and long-term biomarker variations with multi-scale convolutions, and StageNet \cite{gao2020stagenet}, which incorporates disease stage information with a stage-aware LSTM and adaptive convolutional module. Both improve prediction and interpretability but largely operate on structured features without integrating broader knowledge or reasoning capabilities. LINKO \cite{kerdabadi2025multiontologyintegrationdualaxispropagation} uses LLM-initialized embeddings and dual-axis knowledge propagation, vertical within ontologies and horizontal across them, to capture both intra- and cross-ontology relationships. However, it does not fully exploit LLMs’ potential for the related healthcare tasks.

\vspace{0.1cm}
\noindent\textbf{LLM-Based Approaches.}
The growing capabilities of LLMs have motivated new approaches to clinical prediction. GraphCare \cite{jiang2023graphcare} constructs patient-specific knowledge graphs by combining structured knowledge bases with LLM outputs, using a bi-attention augmented GNN to enhance predictions across various predictive tasks. RAM-EHR \cite{xu-etal-2024-ram} applies LLM-powered dense retrieval over multiple knowledge sources to augment patient representations, paired with consistency regularization to improve robustness. Previous works regarding zero-shot prompting \cite{zhu2024prompting} show that prompts incorporating EHR-specific features enable LLMs to make effective predictions in few-shot scenarios. Instruction-based fine-tuning approaches, such as LlamaCare \cite{li-etal-2024-llamacare}, align general-purpose LLMs with clinical vocabulary and tasks, improving quality as judged by human evaluators.
COCONUT (Chain of Continuous Thought) \cite{hao2024training} enables reasoning directly in the LLM’s latent space, exploring multiple inference paths rather than committing to a single chain-of-thought. This reduces premature commitment to a single trajectory and is promising for complex, high-stakes decision-making, such as differential diagnosis or treatment planning. However, current LLM methods still underutilize structured EHR data and fail to jointly model temporal dependencies and hierarchical relationships.

\section{Conclusion}
In this work, we addressed two fundamental limitations that arise when applying Large Language Models to Electronic Health Records: the lack of temporal awareness and the inability to capture patient-to-patient similarity patterns from raw text alone. To overcome these challenges, we introduced {RePrompT}, a time-aware and adaptable framework that integrates structured EHR representations with pretrained LLMs through soft prompt tuning. Experimental results on two large-scale clinical datasets, MIMIC-III and MIMIC-IV, demonstrate that RePrompT consistently outperforms both traditional EHR-based and standard LLM-based baselines.

\section{Limitations}
Despite promising experimental results, several limitations remain. First, the framework relies on the quality of the EHR data; domain shifts in other healthcare systems may affect generalizability. Second, future work is needed to extend the approach to more clinical prediction tasks. 
\section{Ethical Considerations}
A potential risk and ethical consideration of this approach is that using non-anonymized or insufficiently de-identified EHR data may compromise patient privacy, underscoring the need for compliance with relevant data protection regulations.


\bibliography{references}

@article{safedrug,
  title={Safedrug: Dual molecular graph encoders for recommending effective and safe drug combinations},
  author={Yang, Chaoqi and Xiao, Cao and Ma, Fenglong and Glass, Lucas and Sun, Jimeng},
  journal={arXiv preprint arXiv:2105.02711},
  year={2021}
}

@article{retain,
  title={Retain: An interpretable predictive model for healthcare using reverse time attention mechanism},
  author={Choi, Edward and Bahadori, Mohammad Taha and Sun, Jimeng and Kulas, Joshua and Schuetz, Andy and Stewart, Walter},
  journal={Advances in neural information processing systems},
  volume={29},
  year={2016}
}

@article{MIMIC3,
  title={MIMIC-III, a freely accessible critical care database},
  author={Johnson, Alistair EW and Pollard, Tom J and Shen, Lu and Lehman, Li-wei H and Feng, Mengling and Ghassemi, Mohammad and Moody, Benjamin and Szolovits, Peter and Anthony Celi, Leo and Mark, Roger G},
  journal={Scientific data},
  volume={3},
  number={1},
  pages={1--9},
  year={2016},
  publisher={Nature Publishing Group}
}

@inproceedings{gct,
  title={Learning the graphical structure of electronic health records with graph convolutional transformer},
  author={Choi, Edward and Xu, Zhen and Li, Yujia and Dusenberry, Michael and Flores, Gerardo and Xue, Emily and Dai, Andrew},
  booktitle={Proceedings of the AAAI conference on artificial intelligence},
  volume={34},
  pages={606--613},
  year={2020}
}

@article{pytorch,
  title={Pytorch: An imperative style, high-performance deep learning library},
  author={Paszke, Adam and Gross, Sam and Massa, Francisco and Lerer, Adam and Bradbury, James and Chanan, Gregory and Killeen, Trevor and Lin, Zeming and Gimelshein, Natalia and Antiga, Luca and others},
  journal={Advances in neural information processing systems},
  volume={32},
  year={2019}
}

@inproceedings{pyhealth,
    author = {Yang, Chaoqi and Wu, Zhenbang and Jiang, Patrick and Lin, Zhen and Gao, Junyi and Danek, Benjamin and Sun, Jimeng},
    title = {{PyHealth}: A Deep Learning Toolkit for Healthcare Predictive Modeling},
    url = {https://github.com/sunlabuiuc/PyHealth},
    booktitle = {Proceedings of the 27th ACM SIGKDD International Conference on Knowledge Discovery and Data Mining (KDD) 2023},
    year = {2023}
}

@article{jiang2023graphcare,
  title={Graphcare: Enhancing healthcare predictions with personalized knowledge graphs},
  author={Jiang, Pengcheng and Xiao, Cao and Cross, Adam and Sun, Jimeng},
  journal={arXiv preprint arXiv:2305.12788},
  year={2023}
}

@inproceedings{ARCI,
  title={Contrastive learning on medical intents for sequential prescription recommendation},
  author={Hadizadeh Moghaddam, Arya and Nayebi Kerdabadi, Mohsen and Liu, Mei and Yao, Zijun},
  booktitle={Proceedings of the 33rd ACM International Conference on Information and Knowledge Management},
  pages={748--757},
  year={2024}
}

@inproceedings{GRAM,
  title={GRAM: graph-based attention model for healthcare representation learning},
  author={Choi, Edward and Bahadori, Mohammad Taha and Song, Le and Stewart, Walter F and Sun, Jimeng},
  booktitle={Proceedings of the 23rd ACM SIGKDD international conference on knowledge discovery and data mining},
  pages={787--795},
  year={2017}
}

@article{johnson2020mimic,
  title={Mimic-iv},
  author={Johnson, Alistair and Bulgarelli, Lucas and Pollard, Tom and Horng, Steven and Celi, Leo Anthony and Mark, Roger},
  journal={PhysioNet. Available online at: https://physionet. org/content/mimiciv/1.0/(accessed August 23, 2021)},
  pages={49--55},
  year={2020}
}

@article{vaswani2017attention,
  title={Attention is all you need},
  author={Vaswani, A},
  journal={Advances in Neural Information Processing Systems},
  year={2017}
}

@article{johnson2023mimic,
  title={MIMIC-IV, a freely accessible electronic health record dataset},
  author={Johnson, Alistair EW and Bulgarelli, Lucas and Shen, Lu and Gayles, Alvin and Shammout, Ayad and Horng, Steven and Pollard, Tom J and Hao, Sicheng and Moody, Benjamin and Gow, Brian and others},
  journal={Scientific data},
  volume={10},
  number={1},
  pages={1},
  year={2023},
  publisher={Nature Publishing Group UK London}
}

@article{nguyen2016mathtt,
  title={Deepr: a convolutional net for medical records},
  author={Nguyen, Phuoc and Tran, Truyen and Wickramasinghe, Nilmini and Venkatesh, Svetha},
  journal={IEEE journal of biomedical and health informatics},
  volume={21},
  number={1},
  pages={22--30},
  year={2016},
  publisher={IEEE}
}

@article{normalllm2,
  title={Gemini: a family of highly capable multimodal models},
  author={Team, Gemini and Anil, Rohan and Borgeaud, Sebastian and Alayrac, Jean-Baptiste and Yu, Jiahui and Soricut, Radu and Schalkwyk, Johan and Dai, Andrew M and Hauth, Anja and Millican, Katie and others},
  journal={arXiv preprint arXiv:2312.11805},
  year={2023}
}

@inproceedings{healtllm1,
  title={Healai: A healthcare llm for effective medical documentation},
  author={Goyal, Sagar and Rastogi, Eti and Rajagopal, Sree Prasanna and Yuan, Dong and Zhao, Fen and Chintagunta, Jai and Naik, Gautam and Ward, Jeff},
  booktitle={Proceedings of the 17th ACM International Conference on Web Search and Data Mining},
  pages={1167--1168},
  year={2024}
}

@inproceedings{healthllm2,
  title={Llm-based framework for administrative task automation in healthcare},
  author={Gebreab, Senay A and Salah, Khaled and Jayaraman, Raja and ur Rehman, Muhammad Habib and Ellaham, Samer},
  booktitle={2024 12th International Symposium on Digital Forensics and Security (ISDFS)},
  pages={1--7},
  year={2024},
  organization={IEEE}
}

@article{healthllm3,
  title={Prompt engineering as an important emerging skill for medical professionals: tutorial},
  author={Mesk{\'o}, Bertalan},
  journal={Journal of medical Internet research},
  volume={25},
  pages={e50638},
  year={2023},
  publisher={JMIR Publications Toronto, Canada}
}

@article{llmhealth4,
  title={OpenMedLM: prompt engineering can out-perform fine-tuning in medical question-answering with open-source large language models},
  author={Maharjan, Jenish and Garikipati, Anurag and Singh, Navan Preet and Cyrus, Leo and Sharma, Mayank and Ciobanu, Madalina and Barnes, Gina and Thapa, Rahul and Mao, Qingqing and Das, Ritankar},
  journal={Scientific Reports},
  volume={14},
  number={1},
  pages={14156},
  year={2024},
  publisher={Nature Publishing Group UK London}
}

@inproceedings{llmhealth5,
  title={Colacare: Enhancing electronic health record modeling through large language model-driven multi-agent collaboration},
  author={Wang, Zixiang and Zhu, Yinghao and Zhao, Huiya and Zheng, Xiaochen and Sui, Dehao and Wang, Tianlong and Tang, Wen and Wang, Yasha and Harrison, Ewen and Pan, Chengwei and others},
  booktitle={Proceedings of the ACM on Web Conference 2025},
  pages={2250--2261},
  year={2025}
}

@inproceedings{hap,
  title={Hierarchical attention propagation for healthcare representation learning},
  author={Zhang, Muhan and King, Christopher R and Avidan, Michael and Chen, Yixin},
  booktitle={Proceedings of the 26th ACM SIGKDD International Conference on Knowledge Discovery \& Data Mining},
  pages={249--256},
  year={2020}
}

@article{lester2021power,
  title={The power of scale for parameter-efficient prompt tuning},
  author={Lester, Brian and Al-Rfou, Rami and Constant, Noah},
  journal={arXiv preprint arXiv:2104.08691},
  year={2021}
}

@article{wu2023infoprompt,
  title={Infoprompt: Information-theoretic soft prompt tuning for natural language understanding},
  author={Wu, Junda and Yu, Tong and Wang, Rui and Song, Zhao and Zhang, Ruiyi and Zhao, Handong and Lu, Chaochao and Li, Shuai and Henao, Ricardo},
  journal={Advances in neural information processing systems},
  volume={36},
  pages={61060--61084},
  year={2023}
}

@article{healthllm7,
  title={Prompt engineering paradigms for medical applications: scoping review},
  author={Zaghir, Jamil and Naguib, Marco and Bjelogrlic, Mina and N{\'e}v{\'e}ol, Aur{\'e}lie and Tannier, Xavier and Lovis, Christian},
  journal={Journal of Medical Internet Research},
  volume={26},
  pages={e60501},
  year={2024},
  publisher={JMIR Publications Toronto, Canada}
}

@article{vu2021spot,
  title={Spot: Better frozen model adaptation through soft prompt transfer},
  author={Vu, Tu and Lester, Brian and Constant, Noah and Al-Rfou, Rami and Cer, Daniel},
  journal={arXiv preprint arXiv:2110.07904},
  year={2021}
}

@article{liu2024deepseek,
  title={Deepseek-v3 technical report},
  author={Liu, Aixin and Feng, Bei and Xue, Bing and Wang, Bingxuan and Wu, Bochao and Lu, Chengda and Zhao, Chenggang and Deng, Chengqi and Zhang, Chenyu and Ruan, Chong and others},
  journal={arXiv preprint arXiv:2412.19437},
  year={2024}
}

@article{hu2022lora,
  title={Lora: Low-rank adaptation of large language models.},
  author={Hu, Edward J and Shen, Yelong and Wallis, Phillip and Allen-Zhu, Zeyuan and Li, Yuanzhi and Wang, Shean and Wang, Lu and Chen, Weizhu and others},
  journal={ICLR},
  volume={1},
  number={2},
  pages={3},
  year={2022}
}

@article{wang2024calibrating,
  title={Calibrating Verbalized Probabilities for Large Language Models},
  author={Wang, Cheng and Szarvas, Gyuri and Balazs, Georges and Danchenko, Pavel and Ernst, Patrick},
  journal={arXiv preprint arXiv:2410.06707},
  year={2024}
}

@inproceedings{zhang2021grasp,
  title={GRASP: generic framework for health status representation learning based on incorporating knowledge from similar patients},
  author={Zhang, Chaohe and Gao, Xin and Ma, Liantao and Wang, Yasha and Wang, Jiangtao and Tang, Wen},
  booktitle={Proceedings of the AAAI conference on artificial intelligence},
  volume={35},
  pages={715--723},
  year={2021}
}

@inproceedings{ma2020adacare,
  title={Adacare: Explainable clinical health status representation learning via scale-adaptive feature extraction and recalibration},
  author={Ma, Liantao and Gao, Junyi and Wang, Yasha and Zhang, Chaohe and Wang, Jiangtao and Ruan, Wenjie and Tang, Wen and Gao, Xin and Ma, Xinyu},
  booktitle={Proceedings of the AAAI Conference on Artificial Intelligence},
  volume={34},
  pages={825--832},
  year={2020}
}

@inproceedings{gao2020stagenet,
  title={Stagenet: Stage-aware neural networks for health risk prediction},
  author={Gao, Junyi and Xiao, Cao and Wang, Yasha and Tang, Wen and Glass, Lucas M and Sun, Jimeng},
  booktitle={Proceedings of the web conference 2020},
  pages={530--540},
  year={2020}
}

@article{wang2025capabilities,
  title={Capabilities of gpt-5 on multimodal medical reasoning},
  author={Wang, Shansong and Hu, Mingzhe and Li, Qiang and Safari, Mojtaba and Yang, Xiaofeng},
  journal={arXiv preprint arXiv:2508.08224},
  year={2025}
}

@article{hao2024training,
  title={Training large language models to reason in a continuous latent space},
  author={Hao, Shibo and Sukhbaatar, Sainbayar and Su, DiJia and Li, Xian and Hu, Zhiting and Weston, Jason and Tian, Yuandong},
  journal={arXiv preprint arXiv:2412.06769},
  year={2024}
}

@article{zhu2024prompting,
  title={Prompting large language models for zero-shot clinical prediction with structured longitudinal electronic health record data},
  author={Zhu, Yinghao and Wang, Zixiang and Gao, Junyi and Tong, Yuning and An, Jingkun and Liao, Weibin and Harrison, Ewen M and Ma, Liantao and Pan, Chengwei},
  journal={arXiv preprint arXiv:2402.01713},
  year={2024}
}

@inproceedings{xu-etal-2024-ram,
  title = {RAM-EHR: Retrieval Augmentation Meets Clinical Predictions on Electronic Health Records},
  author = {Xu, Ran  and Shi, Wenqi  and Yu, Yue  and Zhuang, Yuchen  and Jin, Bowen  and Wang, May Dongmei  and Ho, Joyce  and Yang, Carl},
  editor = {Ku, Lun-Wei  and Martins, Andre  and Srikumar, Vivek},
  booktitle = {Proceedings of the 62nd Annual Meeting of the Association for Computational Linguistics (Volume 2: Short Papers)},
  month = aug,
  year = "2024",
  address = "Bangkok, Thailand",
  publisher = "Association for Computational Linguistics",
  url = "https://aclanthology.org/2024.acl-short.68/",
  doi = "10.18653/v1/2024.acl-short.68",
  pages = "754--765",
}

@inproceedings{zhou2021informer,
  title={Informer: Beyond efficient transformer for long sequence time-series forecasting},
  author={Zhou, Haoyi and Zhang, Shanghang and Peng, Jieqi and Zhang, Shuai and Li, Jianxin and Xiong, Hui and Zhang, Wancai},
  booktitle={Proceedings of the AAAI conference on artificial intelligence},
  volume={35},
  pages={11106--11115},
  year={2021}
}

@inproceedings{li-etal-2024-llamacare,
  title = {LlamaCare: An Instruction Fine-Tuned Large Language Model for Clinical NLP},
  author = {Li, Rumeng and Wang, Xun and Yu, Hong},
  booktitle = "Proceedings of the 2024 Joint International Conference on Computational Linguistics, Language Resources and Evaluation (LREC-COLING 2024)",
  month = may,
  year = "2024",
  address = "Torino, Italia",
  publisher = "ELRA and ICCL",
  url = "https://aclanthology.org/2024.lrec-main.930/",
  pages = "10632--10641",
}

@misc{kerdabadi2025multiontologyintegrationdualaxispropagation,
      title={Multi-Ontology Integration with Dual-Axis Propagation for Medical Concept Representation}, 
      author={Mohsen Nayebi Kerdabadi and Arya Hadizadeh Moghaddam and Dongjie Wang and Zijun Yao},
      year={2025},
      eprint={2508.21320},
      archivePrefix={arXiv},
      primaryClass={cs.AI},
      url={https://arxiv.org/abs/2508.21320}, 
}

@article{behnamghader2024llm2vec,
  title={Llm2vec: Large language models are secretly powerful text encoders},
  author={BehnamGhader, Parishad and Adlakha, Vaibhav and Mosbach, Marius and Bahdanau, Dzmitry and Chapados, Nicolas and Reddy, Siva},
  journal={arXiv preprint arXiv:2404.05961},
  year={2024}
}

@article{wolf2019huggingface,
  title={Huggingface's transformers: State-of-the-art natural language processing},
  author={Wolf, Thomas and Debut, Lysandre and Sanh, Victor and Chaumond, Julien and Delangue, Clement and Moi, Anthony and Cistac, Pierric and Rault, Tim and Louf, R{\'e}mi and Funtowicz, Morgan and others},
  journal={arXiv preprint arXiv:1910.03771},
  year={2019}
}

@article{loshchilov2017decoupled,
  title={Decoupled weight decay regularization},
  author={Loshchilov, Ilya and Hutter, Frank},
  journal={arXiv preprint arXiv:1711.05101},
  year={2017}
}

@article{cheong2023adaptive,
  title={Adaptive integration of categorical and multi-relational ontologies with ehr data for medical concept embedding},
  author={Cheong, Chin Wang and Yin, Kejing and Cheung, William K and Fung, Benjamin CM and Poon, Jonathan},
  journal={ACM Transactions on Intelligent Systems and Technology},
  volume={14},
  number={6},
  pages={1--20},
  year={2023},
  publisher={ACM New York, NY}
}

@inproceedings{gao2023small,
  title={Small pre-trained language models can be fine-tuned as large models via over-parameterization},
  author={Gao, Ze-Feng and Zhou, Kun and Liu, Peiyu and Zhao, Wayne Xin and Wen, Ji-Rong},
  booktitle={Proceedings of the 61st Annual Meeting of the Association for Computational Linguistics (Volume 1: Long Papers)},
  pages={3819--3834},
  year={2023}
}

@article{bucher2024fine,
  title={Fine-tuned'small'LLMs (still) significantly outperform zero-shot generative AI models in text classification},
  author={Bucher, Martin Juan Jos{\'e} and Martini, Marco},
  journal={arXiv preprint arXiv:2406.08660},
  year={2024}
}

@article{bozkurt2016contributory,
  title={Contributory risk and management of comorbidities of hypertension, obesity, diabetes mellitus, hyperlipidemia, and metabolic syndrome in chronic heart failure: a scientific statement from the American Heart Association},
  author={Bozkurt, Biykem and Aguilar, David and Deswal, Anita and Dunbar, Sandra B and Francis, Gary S and Horwich, Tamara and Jessup, Mariell and Kosiborod, Mikhail and Pritchett, Allison M and Ramasubbu, Kumudha and others},
  journal={Circulation},
  volume={134},
  number={23},
  pages={e535--e578},
  year={2016},
  publisher={Lippincott Williams \& Wilkins Hagerstown, MD}
}

@article{tan2024language,
  title={Are language models actually useful for time series forecasting?},
  author={Tan, Mingtian and Merrill, Mike and Gupta, Vinayak and Althoff, Tim and Hartvigsen, Tom},
  journal={Advances in Neural Information Processing Systems},
  volume={37},
  pages={60162--60191},
  year={2024}
}

@article{liu2019hierarchical,
  title={Hierarchical transformers for multi-document summarization},
  author={Liu, Yang and Lapata, Mirella},
  journal={arXiv preprint arXiv:1905.13164},
  year={2019}
}
\appendix
\newpage
\section{Appendix}

\begin{table}[ht]
\centering
\begin{tabular}{l c}
\hline
\textbf{Model} & \textbf{Time (s)} \\
\hline
Deepr     & 0.08 \\
RETAIN    & 0.05 \\
GRASP     & 0.14 \\
AdaCare   & 0.03 \\
GRAM      & 0.17 \\
StageNet  & 0.07 \\
Adore     & 0.19 \\
ARCI      & 0.94 \\
RePrompT  & 1.44 \\
\hline
\end{tabular}
\caption{Computational time analysis of the proposed method when processing a batch of eight patients.}
\vspace{-15pt}
\label{tb:comp}
\end{table}

\subsection{RETAIN Encoder Details}
\label{app:retain}

For the Struct-Encoded Prompt Tuning we adopt RETAIN \cite{retain}, a structured EHR encoder that summarizes
sequential medical codes into dense patient embeddings using a dual-level
attention mechanism.

Formally, RETAIN employs two sets of attention weights: visit-level attention
$\{\bm{\alpha}_{i,j}\}_{j=1}^{t}$ and variable-level attention
$\{\bm{\beta}_{i,j}\}_{j=1}^{t}$. Visit-level attention determines the relative
importance of each visit embedding $\{\boldsymbol{V}_{i,j}\}_{j=1}^{t}$:
\begin{equation}
\label{eq:att_1}
\{\boldsymbol{g}_{i,j}\}_{j=1}^{T_i} =
\mathrm{GRU}_{\alpha}(\{\boldsymbol{V}_{i,j}\}_{j=1}^{T_i})
\end{equation}

\begin{equation}
\label{eq:softmax}
\{\bm{\alpha}_{i,j}\}_{j=1}^{t} =
\mathrm{Softmax}(\boldsymbol{w}_{\alpha} \{\boldsymbol{g}_{i,j}\}_{j=1}^{t} +
b_{\alpha})
\end{equation}

Variable-level attention highlights the contribution of individual medical codes
within each visit:

\begin{equation}
\label{eq:att_2}
\{\boldsymbol{h}_{i,j}\}_{j=1}^{t} =
\mathrm{GRU}_{\beta}(\{\boldsymbol{V}_{i,j}\}_{j=1}^{t})
\end{equation}

\begin{equation}
\label{eq:tanh}
\{\bm{\beta}_{i,j}\}_{j=1}^{t} =
\tanh(\boldsymbol{w}_{\beta}\{\boldsymbol{h}_{i,j}\}_{j=1}^{t} + b_{\beta})
\end{equation}
Final patient representation is computed based on both attention values:
\begin{equation}
\label{eq:attention_dots}
\boldsymbol{k}_{i,t} =
\sum_{j=1}^{t} \bm{\alpha}_{i,j} \bm{\beta}_{i,j} \odot \boldsymbol{V}_{i,j}
\end{equation}

\begin{equation}
\label{eq:enc}
\boldsymbol{S}_{i,t} =
\boldsymbol{w}_{\mathrm{enc}} \boldsymbol{k}_{i,t} + b_{\mathrm{enc}}
\end{equation}

In the equations, $\boldsymbol{w}_{\alpha}$, $b_{\alpha}$, $\boldsymbol{w}_{\beta}$,
$b_{\beta}$, $\boldsymbol{w}_{\mathrm{enc}}$, and $b_{\mathrm{enc}}$ are trainable
parameters. The resulting embedding $\boldsymbol{S}_{i,t}$ is used as a
structured soft prompt for the LLM.
\subsection{Computational Time Analysis}
\label{app:comp_ta}
Table \ref{tb:comp} reports the inference time required to generate predictions for a batch of eight patients. Although our model incurs higher latency than conventional deep learning baselines, the overall inference time remains practical for real-world deployment and is still fast enough to support timely prediction in realistic clinical settings.

\subsection{Performance Comparison on Medication Recommendation Task}
\label{app:comp_med}
Table \ref{tab:medication_results} presents the medication recommendation results on the MIMIC-III and MIMIC-IV datasets. Note that for medication recommendation, we do not group medications according to the ATC ontology, as our goal is to directly recommend specific medications rather than medication classes.  RePrompT shows strong and competitive performance against the baseline methods across the reported metrics. These results show that integrating patient-specific EHR embeddings with the LLM through time-aware prompting provides useful clinical context for multi-label medication recommendation.

\begin{table}[ht]
\centering
\renewcommand{\arraystretch}{1.1}
\footnotesize
\setlength{\tabcolsep}{4pt}
\begin{tabular}{l|cc|cc}
\hline
\multirow{2}{*}{\textbf{Model}}
& \multicolumn{2}{c|}{\textbf{MIMIC-IV}}
& \multicolumn{2}{c}{\textbf{MIMIC-III}} \\ \cline{2-5}
& F1 & Jaccard & F1 & Jaccard \\ \hline

Deepr
& 0.222 & 0.184 & 0.249 & 0.156 \\

RETAIN
& \uline{0.344} & \uline{0.251} & 0.251 & 0.158 \\

GRAM
& 0.236 & 0.168 & 0.197 & 0.122 \\

GRASP
& 0.264 & 0.189 & 0.150 & 0.093 \\

AdaCare
& 0.244 & 0.175 & 0.205 & 0.129 \\

StageNet
& 0.270 & 0.194 & 0.236 & 0.137 \\

Adore
& 0.201 & 0.159 & 0.147 & 0.123 \\

ARCI
& 0.308 & 0.245 & \uline{0.285} & \uline{0.187} \\

\textbf{RePrompT}
& \textbf{0.374} & \textbf{0.276} & \textbf{0.314} & \textbf{0.204} \\

\hline
\end{tabular}
\caption{Performance comparison on the medication recommendation task on the MIMIC-IV and MIMIC-III datasets, evaluated by F1-score and Jaccard similarity.}

\label{tab:medication_results}
\end{table}

\end{document}